\documentclass[sigconf,fleqn]{article}

\usepackage{booktabs} % For formal tables
\usepackage{subcaption}
\usepackage{multirow}
\usepackage{graphicx}
\usepackage{rotating}

\newcommand{\eg}{{\it e.g.}}
\newcommand{\ie}{{\it i.e.}}
\newcommand{\etal}{{\it et al. }}

\begin{document}
\title{Accelerating the Evolution of Convolutional Neural Networks with Node-Level Mutations and Epigenetic Weight Initialization}

\author{Travis Desell\thanks{Rochester Institute of Technology. 134 Lomb Memorial Drive, Rochester, NY 14623.}}

\maketitle

\begin{abstract}
    This paper examines three generic strategies for improving the performance of neuro-evolution techniques aimed at evolving convolutional neural networks (CNNs). These were implemented as part of the Evolutionary eXploration of Augmenting Convolutional Topologies (EXACT) algorithm. EXACT evolves arbitrary convolutional neural networks (CNNs) with goals of better discovering and understanding new effective architectures of CNNs for machine learning tasks and to potentially automate the process of network design and selection.  The strategies examined are node-level mutation operations, epigenetic weight initialization and pooling connections.   Results were gathered over the period of a month using a volunteer computing project, where over 225,000 CNNs were trained and evaluated across 16 different EXACT searches. The node mutation operations where shown to dramatically improve evolution rates over traditional edge mutation operations (as used by the NEAT algorithm), and epigenetic weight initialization was shown to further increase the accuracy and generalizability of the trained CNNs. As a negative but interesting result, allowing for pooling connections was shown to degrade the evolution progress. The best trained CNNs reached 99.46\% accuracy on the MNIST test data in under 13,500 CNN evaluations -- accuracy comparable with some of the best human designed CNNs.   

\end{abstract}

\section{Introduction}
\label{sec:introduction}

Convolutional Neural Networks (CNNs) have become a highly active area of research due to strong results in areas such as image classification~\cite{krizhevsky2012imagenet,lecun1998gradient}, video classification~\cite{karpathy2014large}, sentence classification~\cite{kim2014convolutional}, and speech recognition~\cite{hinton2012deep}, among others. Significant progress has been made in the design of CNNs, from the venerable LeNet 5~\cite{lecun1998gradient} to more recent large and deep networks such as AlexNet~\cite{krizhevsky2012imagenet}, VGGNet~\cite{simonyan2014very}, GoogleNet~\cite{szegedy2015going} and ResNet~\cite{he2016deep}. However, less work has been made in the area of automated design of CNNs. By investigating strategies for accelerating the neuro-evolution process, this work provides advances towards answering a big open challenge in machine learning: {\it What is the optimal architecture for a convolutional neural network?} 

Three strategies are investigated: {\it i)} {\it node-level mutation operators} which allow for faster growth of the evolved CNNs; {\it ii)} {\it epigenetic weight initialization} where child CNNs retain weights from their already trained parent(s); and {\it iii)} {\it fractional max pooling}~\cite{graham2014fractional}, which allows for the use of pooling connections between arbitrary sized input and output feature maps.

While generically applicable to many neuro-evolution algorithms, these strategies were implemented as part of the Evolutionary eXploration of Augmenting Convolutional Topologies (EXACT), which can evolve CNNs of arbitrary structure, filter and feature map size. Due to high computational demands, it has been implemented as part of the Citizen Science Grid\footnote{http://csgrid.org}, a Berkeley Open Infrastructure for Network Computing (BOINC)~\cite{anderson_boinc_2005} volunteer computing project. 

This allowed 13,000 volunteered CPUs to train over 225,000 CNNs in the period of a month, with the best evolved CNN reaching 99.46\% accuracy on the MNIST handwritten digits dataset~\cite{lecun1998mnist}. Interestingly, allowing pooling edges degraded search performance, most likely due to increasing the search space. However, epigenetic weight initialization was shown to provide a significant improvement to test error rates, which were reduced by 0.11\% to 0.44\%, and on average by 0.20\% to 0.62\% when compared to weights initialized by the standard randomized method proposed by He \etal~\cite{he2015delving}. Node mutations reduced test error by 0.21\% for the best found genomes and on average 0.175\% for the searches without pooling and by 0.18\% for the best found genomes and by 0.1575\% for the searches with pooling.  These results are significant as error rates were already below 1\%.

\section{Related Work}
\label{sec:related_work}

There exist a number of neuro-evolution techniques capable of evolving the structure of feed forward and recurrent neural networks, such as NEAT~\cite{stanley2002evolving}, HyperNEAT~\cite{stanley2009hypercube}, CoSyNE~\cite{gomez2008accelerated}, as well as ant colony optimization based approaches~\cite{desell-evostar-2015,salama2014novel}. However, these have not yet been applied to CNNs due to the size and structure of CNNs, not to mention the significant amount of time required to train one.

Zoph \etal~\cite{zoph2016neural} propose a method which uses a recurrent neural network trained with reinforcement learning to maximize the expected accuracy of generated architectures on a validation set of images.  However, this approach is gradient based and generates CNNs by layer, with each layer having a fixed filter width and height, stride width and height, and number of filters.

Xie \etal~propose a Genetic CNN method which encodes CNNs as binary strings~\cite{xie2017geneticcnn}, however they only evolve structure of convolution operations between pooling layers, and keep the filter sizes fixed. Suganuma \etal have proposed a method based on Cartesian genetic programming which defines highly functional modules, such as convolutional blocks and tensor concatenation, evolving CNNs with traditional structures~\cite{suganuma2017genetic}. Sun \etal have presented a similar method which utilizes a variable length gene encoding strategy to evolve structured CNNs~\cite{sun-arxiv-evocnn-2017}. Miikkulainen \etal ~have proposed CoDeepNEAT, which is also based on NEAT with each node acting as an entire layer, with the type of layer and hyperparameters being co-evolved~\cite{mikkulainen2017codeepneat}. However, connections within layers are fixed depending on their type without arbitrary connections.   Real \etal have evolved image classifiers on the CIFAR-10 and CIFAR-100 datasets~\cite{real2017evolution}. They use a distributed algorithm to evolve progressively more complex CNNs through mutation operations, and handle conflicts in filter sizes by reshaping non-primary edges with zeroth order interpolation. In their case, mutation operators work on entire layers.

To our knowledge, the EXACT algorithm is the only method which can evolve completely arbitrary CNN structures. This allows for CNNs to be evolved without any explicit layering (as can be seen in Figure~\ref{fig:example_genomes}). This allows for the potential to discover new substructures and architecture types that aren't bounded by traditional structural confines, and more fine-tuned refinement of the overall CNN architecture. Further, we are unaware of other strategies for evolving neural networks that utilize epigenetic weight initialization or the proposed node-level mutations.

%modern neruo evolution: Wierstra et al. (2005); Floreano et al. (2008); Stanley et al. (2009)

%Daan Wierstra, Faustino J Gomez, and Jurgen Schmidhuber. Modeling systems with internal state using evolino. In GECCO, 2005.

%Dario Floreano, Peter Durr, and Claudio Mattiussi. Neuroevolution: from architectures to learning. Evolutionary Intelligence, 2008.

%Kenneth O. Stanley, David B. D’Ambrosio, and Jason Gauci. A hypercube-based encoding for evolving large-scale neural networks. Artificial Life, 2009.

\section{Evolutionary Exploration of Augmenting Convolutional Topologies}
\label{sec:exact}

The EXACT algorithm starts with the observation that any two feature maps of any size within a CNN can be connected by a convolution of size $conv_d = |out_d - in_d| + 1$, where $out_d$ and $in_d$ are the size of the output and input feature maps, respectively, and $conv_d$ is the size of the convolution in dimension $d$. The consequence of this observation is that the structure of a CNN can be evolved solely by determining the sizes of the feature maps and how they are connected. Instead of evolving the weights of individual neurons and how they are connected, as done in the NEAT~\cite{stanley2002evolving} algorithm, the architecture of a CNN can be evolved in a similar fashion except on the level of how feature maps are connected, with additional operators to modify the feature map sizes.  Whereas NEAT works on the level of neurons and weights, EXACT works on the level of feature maps (or \emph{nodes}) and filters (or \emph{edges}).

Due to the computational expense of training CNNs, EXACT has been designed with scalable distributed execution in mind. It uses an asynchronous evolution strategy, which has been shown by Desell \etal ~to allow scalability up to potentially millions of compute hosts in a manner independent of population size~\cite{desell-analysis-massive-eas-2010}. A master process manages a population of \emph{CNN genomes} (the feature map sizes and how they are connected) along with their fitness (the minimized error after backpropagation on that CNN). Worker processes request CNN genomes to evaluate from the master, which generates them either through applying mutation operations to a randomly selected genome in the population or by selecting two parents and performing crossover to generate a child genome.  When that worker completes training the CNN, it reports the CNN along with its fitness back to the master, which will insert it into the population and remove the least fit genome if it would improve the population.  This asynchronous approach has an additional benefit that no worker need wait for the results of another worker to request another CNN to evaluate, which is particularly important in that the evolved CNNs have different training times -- this approach automatically load balances itself. Due to space limitations, the reader is referred to Desell~\cite{desell2017developing} for further implementation details.

Given the strong advances made by the machine learning community for training deep CNNs and the sheer number of weights in large CNNs, attempts to try and simultaneously evolve the weights in the neural networks did not seem feasible.  Instead, EXACT allows for any CNN training method to be plugged in to perform the fitness evaluation done by the workers.  In this way, EXACT can benefit from further advances by the machine learning community and also make for an interesting platform to evaluate different neural network training algorithms.

\subsection{Evolving Networks for Generalizability}
\label{sec:evolving_for_generalizability}

To prevent biasing the structure of the evolved CNNs to the test data the training data was separated into two randomly selected subsets. In the case of MNIST, 50,000 of the 60,000 training images were used to train the neural networks, and the other 10,000 images were used as a validation set. The fitness of an evolved CNN was the error (cross entropy loss) on the validation data.  CNNs were only evaluated on the test data to generate final results, the test data was never utilized in either the training of CNNs or in the evolutionary process of EXACT.  This ensured that the CNNs evolved were actually generalizable, in fact, as shown in Table~\ref{table:search_rates}, they tended to over perform on the test data as compared to the validation set.

\subsection{Population Initialization}
\label{sec:population_initialization}

Generation of the initial population starts with first generating a \emph{minimal CNN genome}, which consists solely of the input node, which is the size of the training images (plus padding if desired), and one output node per training class for a softmax output layer, with one edge connecting the input node to each output node.  In this case of this work which uses the MNIST handwritten digits dataset, this is a 28x28 input node, and 10 1x1 output nodes. This is sent as the CNN genome for the first work request, and a copy of it is inserted into the population with $\infty$ as fitness, denoting that it had not been evaluated yet. Further work requests are fulfilled by taking a random member of the population (which will be initially just the minimal CNN genome), mutating it, inserting a copy of the mutation into the population with $\infty$ as fitness and sending that mutated CNN genome to the worker to evaluate.  Once the population has reached a user specified population size through inserting newly generated mutations and results received by workers, work requests are fulfilled by either mutation or crossover, depending on a user specified crossover rate (\eg, a 20\% crossover rate will result in 80\% mutation).

\subsection{Epigenetic Weight Initialization}
\label{sec:epigenetic_weights}

For CNNs generated during population initialization, the weights are initialized as recommended by He \etal~\cite{he2015delving}, where the variance, $\sigma^2$, of the weights, $w$, input to a neuron is $\sigma^2(w) = \sqrt{\frac{2}{n}}$, where $n$ is the number of weights input to that neuron. However, after the initial population is evaluated, child genomes are generated from one or two trained parent CNNs. This work investigates having the weights of the parent CNN genomes be carried over into child genomes, \ie, \emph{"epigenetic" weight initialization} -- these weights are a modification of how the genome is expressed as opposed to a modification of the genome itself.

\subsection{Fractional Max Pooling}
\label{sec:fractional_max_pooling}

Other work with the EXACT algorithm only utilized convolutional edges as performing standard max pooling requires the size of the input feature map to be an integer multiple of the size of the output feature map~\cite{desell2017developing,desell2017large}. Graham's fractional max pooling~\cite{graham2014fractional} provides a solution to this challenge, as it allows for pooling operations between arbitrarily sized feature maps. As an example, if the input feature map is 14x11 and the output feature map is 4x5, then the pooling sizes in the x dimension are 3, 3, 3, 2 and the pooling sizes in the y dimension are 2, 2, 2, 2, 2, 1. Every forward pass through the network the order of these pooling dimensions is randomized, causing varying rectangular pools, which provides beneficial distortions and regularization. Pooling sizes can also be increased to allow for overlap.

As EXACT evolves arbitrary structures, it is possible for multiple pooling edges to be input to a node. In order to appropriately back propagate error, each pooling edge is also given a single weight as a scaling factor. The output of the pooling operation is multiplied by this value. This allows error to be backpropagated on each pooling input according to the scaling factor.

\subsection{Mutation and Recombination Operations}
\label{sec:mutation}

%When a CNN genome is selected for mutation, a user specified number of the following mutations are performed. In testing, this was found to be beneficial to allow for greater variation in the CNNs generated. Each operator is selected with a user specified rate. 

This work investigates edge mutations (as in NEAT), and new high level node mutations. Other operations involve changing the size of a node's feature map and crossover. Whereas NEAT only requires innovation numbers for new edges, EXACT requires innovation numbers for both new nodes and new edges. The master process keeps track of all node and edge innovations made, which is required to perform the crossover operation in linear time without a graph matching algorithm.

\subsubsection{Edge Mutations:}
\label{sec:edge_mutations}

\paragraph{Disable Edge} This operation randomly selects an enabled edge in a CNN genome and disables it so that it is not used.  The edge remains in the genome. As the \emph{disable edge} operation can potentially make an output node unreachable, after all mutation operations have been performed to generate a child CNN genome, if any output node is unreachable that CNN genome is discarded and a new child is generation by another attempt at mutation.

\paragraph{Enable Edge} If there are any disabled edges in the CNN genome, this operation selects a disabled edge at random and enables it.

\paragraph{Split Edge} This operation selects an enabled edge at random and disables it. It creates a new node (creating a new node innovation) and two new edges (creating two new edge innovations), and connects the input node of the split edge to the new node, and the new node to the output node of the split edge. The feature map size of the new node is set to $\lceil \frac{i_x + o_x}{2} \rceil$ by $\lceil \frac{i_y + o_y}{2} \rceil$, where $i_d$ and $o_d$ are the size of the input and output feature maps, respectively, in dimension $d$ (\ie, the size of the new node is halfway between the size of the input and output nodes).  Further, the new node is given a depth value, $depth_{new} = \frac{depth_{output} + depth_{input}}{2.0}$, which is used by the \emph{add edge} operation and to linearly perform forward and backward propagation without graph traversal. If fractional max pooling is being utilized, then the edge types are determined randomly with a 50\% chance for pooling and 50\% chance for convolution.

\paragraph{Add Edge} This operation selects two nodes $n_1$ and $n_2$ within the CNN Genome at random, such that $depth_{n_1} < depth_{n_2}$ and such that there is not already an edge between those nodes in this CNN Genome, and then adds an edge from $n_1$ to $n_2$. This ensures that all edges generated feed forward (currently EXACT does not evolve recurrent CNNs). If an edge between $n_1$ and $n_2$ exists within the master's innovation list, that edge innovation is used, otherwise this creates a new edge innovation. If fractional max pooling is being utilized, then the edge type is determined randomly with a 50\% chance for pooling and 50\% chance for convolution.

\paragraph{Alter Edge Type} This operation is only used if pooling is allowed. This selects an enabled edge at random and changes it from pooling to convolutional or convolutional to pooling.

\subsubsection{Node Mutations:}
\label{sec:node_mutations}

\paragraph{Disable Node} This operation selects a random non-input and non-output node and disabled it along with all of its incoming and outgoing edges.

\paragraph{Enable Node} This operation selects a random disabled node and enables it along with all of its incoming and outgoing edges.

\paragraph{Add Node} This operation selects a random depth between 0 and 1, noninclusive. Given that the input node is always depth 0 and the output nodes are always depth 1, this splits the CNN in two. A new node is created, at that depth, and 1-5 edges are randomly generated to nodes with a lesser depth, and 1-5 edges are randomly generated to nodes with a greater depth.  The node size is set to the average of the maximum input node size and minimum output node size. If fractional max pooling is being used, all input edges have a 50\% chance of being pooling or convolutional, and all output edges have a 50\% chance of being pooling or convolutional.

\paragraph{Split Node} This operation takes one non-input, non-output node at random and splits it. This node is disabled (as in the disable node operation) and two new nodes are created at the same depth as their parent. One input and one output edge are assigned to each of the new nodes, with the others being assigned randomly, ensuring that the newly created nodes have both inputs and outputs. If there is only one input or one output edge to this node, then those edges are duplicated for the new nodes. If fractional max pooling is being utilized, the newly created edges are of the same type as on the node they were split from.

\paragraph{Merge Node} This operation takes two non-input, non-output nodes at random and combines them.  The selected nodes are disabled (as in the disable node operation) and a new node is created with a depth equal to average of its parents. This node is connected to the inputs and outputs of its parents, with input edges created to those with a lower depth, and output edges created to those with a deeper depth. If fractional max pooling is being utilized, the newly created edges are of the same type as on the node they were merged from.

\subsubsection{Other Operations:}
\label{sec:other_operations}

\paragraph{Change Node Size} This operation selects a node at random from within the CNN Genome and randomly increases or decreases its feature map size in both the x and y dimension. For this work, the potential size modifications used were [-2, -1, +1, +2].

\paragraph{Change Node Size X} This operation is the same as \emph{change node size} except that it only changes the feature map size in the x dimension.

\paragraph{Change Node Size Y} This operation is the same as \emph{change node size} except that it only changes the feature map size in the y dimension.

\paragraph{Crossover} Crossover utilizes two hyperparameters, the \emph{more fit parent crossover rate} and the \emph{less fit parent crossover rate}. Two parent CNN genomes are selected, and the child CNN genome is generated from every edge that appears in both parents. Edges that only appear in the more fit parent are added randomly at the \emph{more fit parent crossover rate}, and edges that only appear in the less fit parent are added randomly at the \emph{less fit parent crossover rate}. Edges not added by either parent are also carried over into the child CNN genome, however they are set to disabled. Nodes are then added for each input and output of an edge.  If the more fit parent has a node with the same innovation number, it is added from the more fit parent (\ie, feature map sizes are inherited from the more fit parent if possible), and from the less fit parent otherwise. If fractional max pooling is being utilized, edge types remain the same as in the parent they were inherited from.

\section{Backpropagation and EXACT Hyperparameters}
\label{sec:implementation}

\paragraph{Backpropagation Hyperparameters} A fairly standard implementation of stochastic backpropagation was used to train the CNNs. Backpropagation was done using batch normalization~\cite{ioffe2015batch} with a batch size of 50 and $\alpha = 0.1$. Apart from the softmax output layer, each node in the CNNs evolved by EXACT used a leaky ReLU activation function with max value 5.5 and a leak of 0.1. Weights were updated using Nesterov momentum and L2 Regularization~\cite{bengio2013advances}. Momentum $\mu = 0.5$, $\Delta\mu = 0.95$, learning rate $\eta = 0.0125$, $\Delta\eta = 0.95$, and weight decay $\lambda = 0.0005$, $\Delta\lambda = 0.95$ were used due to consistent good performance in previous work. These parameters were updated every epoch by their $\delta$ values as follows:

\begin{equation}
\mu = \mu_{max} - ((\mu_{max} - \mu) * \Delta\mu)
\end{equation}
\begin{equation}
\eta = max(\eta * \Delta\eta, \eta_{min})
\end{equation}
\begin{equation}
\lambda = max(\lambda * \Delta\lambda, \lambda_{min})
\end{equation}

\paragraph{EXACT Hyperparameters} All searches had a population size of 50, a crossover rate of 20\% (which entails a mutation rate of 80\%) and one mutation operation was performed for each genome mutated. For searches without node operations, mutations were done at the following rates: $\frac{2.5}{15}$ {\emph edge disable}, $\frac{2.5}{15}$ {\emph edge enable}, $\frac{3}{15}$ {\emph edge split}, $\frac{3}{15}$ {\emph edge add}, $\frac{2}{15}$ {\emph node change size}, $\frac{1}{15}$ {\emph node change size x}, and $\frac{1}{15}$ {\emph node change size y}.  For searches with node operations, mutations were done at the following rates: $\frac{2.5}{25}$ {\emph edge disable}, $\frac{2.5}{25}$ {\emph edge enable}, $\frac{3}{25}$ {\emph edge split}, $\frac{3}{25}$ {\emph edge add}, $\frac{2}{25}$ {\emph node change size}, $\frac{1}{25}$ {\emph node change size x}, $\frac{1}{25}$ {\emph node change size y}, $\frac{3}{25}$ {\emph node add}, $\frac{2}{25}$ {\emph node split}, $\frac{2}{25}$ {\emph node merge}, $\frac{1.5}{25}$ {\emph node disable}, and $\frac{1.5}{25}$ {\emph node enable}.

\section{Results}
\label{sec:results}

To examine the effect of pooling and the new node level mutation operations, four different types of EXACT searches were run for a period of a month using a volunteer computing project. The different search types were: {\it i)} node and edge operations, without pooling; {\it ii)} node and edge operations, with pooling; {\it iii)} only edge operations, without pooling; {\it iv)} only edge operations, with pooling.  During this time over 3500 volunteers provided over 13,000 compute hosts to train the evolved CNNs. All 16 searches were run simultaneously to minimize external influences such as the availability of volunteered hosts and network outages. Each search evaluated between 13,000 and 14,500 different trained CNNs. In total over 225,000 CNNs were trained and evaluated during this time period for this work. 

\subsection{Effects of Node Mutations and Pooling}
Table~\ref{table:search_rates} presents the best, average, and worst validation and testing error and prediction rates for each of these searches. The searches with node mutations and no pooling performed the best, with significant improvements over those without node mutations and with pooling. In terms of error and prediction rates, the searches with node mutations and pooling performed comparably to the searches without pooling and without node mutations. The searches with pooling and without node operations performed the worst.

\begin{sidewaystable}
\centering
\begin{small}
\begin{tabular}{|l|r|r|r|r|r|r|r|r|r|r|r|r|}
\hline
                    & \multicolumn{3}{|c|}{{\bf Validation Error}}        & \multicolumn{3}{|c|}{{\bf Testing Error}}         & \multicolumn{3}{|c|}{{\bf Validation Predictions}} & \multicolumn{3}{|c|}{{\bf Testing Predictions}} \\
\cline{2-13}
{\bf Search}       & {\bf Worst}    & {\bf Avg} & {\bf Best}       & {\bf Worst}        & {\bf Avg} & {\bf Best}   & {\bf Worst}    & {\bf Avg} & {\bf Best}       & {\bf Worst}        & {\bf Avg} & {\bf Best}  \\
\hline
\hline
   Node+Edge 1& 259.49 & 234.71 & 224.37 & 245.48 & 220.99 & 188.62 & 99.20\% & 99.28\% & 99.36\% & 99.14\% & 99.26\% & 99.46\% \\
   \hline
   Node+Edge 2& 300.00 & 276.16 & 258.75 & 296.74 & 267.02 & 243.16 & 99.07\% & 99.17\% & 99.27\% & 99.03\% & 99.14\% & 99.30\% \\
   \hline
   Node+Edge 3& 262.86 & 240.13 & 232.08 & 249.33 & 222.21 & 189.20 & 99.20\% & 99.27\% & 99.37\% & 99.13\% & 99.25\% & 99.39\% \\
   \hline
   Node+Edge 4& 281.66 & 256.58 & 247.89 & 273.01 & 235.60 & 210.81 & 99.08\% & 99.19\% & 99.28\% & 99.04\% & 99.21\% & 99.36\% \\
   \hline
   \hline
   Node+Edge, Pooling 1& 346.27 & 320.52 & 302.79 & 359.28 & 326.21 & 274.14 & 98.89\% & 99.02\% & 99.12\% & 98.82\% & 99.00\% & 99.20\% \\
   \hline
   Node+Edge, Pooling 2& 335.53 & 296.80 & 283.96 & 350.83 & 299.32 & 271.21 & 98.95\% & 99.07\% & 99.18\% & 98.82\% & 99.03\% & 99.25\% \\
   \hline
   Node+Edge, Pooling 3& 349.16 & 335.53 & 316.00 & 441.07 & 358.78 & 312.25 & 98.83\% & 98.96\% & 99.11\% & 98.74\% & 98.92\% & 99.15\% \\
   \hline
   Node+Edge, Pooling 4& 357.96 & 333.84 & 323.10 & 372.10 & 320.63 & 284.59 & 98.85\% & 98.96\% & 99.12\% & 98.79\% & 98.96\% & 99.17\% \\
   \hline
   \hline
   Edge 1& 324.57 & 294.86 & 278.76 & 329.86 & 294.91 & 254.20 & 99.01\% & 99.09\% & 99.18\% & 98.86\% & 99.03\% & 99.18\% \\
   \hline
   Edge 2& 321.82 & 308.53 & 298.78 & 363.19 & 319.48 & 280.03 & 98.97\% & 99.07\% & 99.17\% & 98.79\% & 98.98\% & 99.18\% \\
   \hline
   Edge 3& 292.04 & 286.42 & 271.55 & 339.46 & 282.48 & 225.64 & 99.07\% & 99.13\% & 99.22\% & 98.95\% & 99.07\% & 99.26\% \\
   \hline
   Edge 4& 327.44 & 293.25 & 277.72 & 327.98 & 292.53 & 254.06 & 98.97\% & 99.09\% & 99.22\% & 98.89\% & 99.03\% & 99.19\% \\
   \hline
   \hline
   Edge, Pooling 1& 390.09 & 368.79 & 350.83 & 417.36 & 376.38 & 331.14 & 98.76\% & 98.87\% & 99.02\% & 98.65\% & 98.79\% & 99.07\% \\
   \hline
   Edge, Pooling 2& 378.39 & 359.22 & 338.24 & 449.45 & 394.67 & 329.94 & 98.79\% & 98.86\% & 98.95\% & 98.49\% & 98.72\% & 98.98\% \\
   \hline
   Edge, Pooling 3& 389.28 & 354.73 & 327.78 & 421.46 & 359.62 & 314.51 & 98.79\% & 98.92\% & 99.02\% & 98.67\% & 98.82\% & 98.99\% \\
   \hline
   Edge, Pooling 4& 392.37 & 382.92 & 342.80 & 478.48 & 383.83 & 310.83 & 98.67\% & 98.79\% & 98.95\% & 98.56\% & 98.79\% & 99.08\% \\
\hline
\end{tabular}
\caption{\label{table:search_rates} EXACT Search Error and Prediction Rates}
\end{small}
\end{sidewaystable}

\begin{sidewaystable}
\begin{small}
\begin{center}
\begin{tabular}{|l|r|r|r|r|r|r|r|r|r|r|r|r|}
\hline
                    & \multicolumn{3}{|c|}{{\bf Validation Error}}        & \multicolumn{3}{|c|}{{\bf Testing Error}}         & \multicolumn{3}{|c|}{{\bf Validation Predictions}} & \multicolumn{3}{|c|}{{\bf Testing Predictions}} \\
\cline{2-13}
{\bf Search}       & {\bf Worst}    & {\bf Avg} & {\bf Best}       & {\bf Worst}        & {\bf Avg} & {\bf Best}   & {\bf Worst}    & {\bf Avg} & {\bf Best}       & {\bf Worst}        & {\bf Avg} & {\bf Best}  \\
\hline
\hline
    Node+Edge           & 313.83 & 291.71 & 268.00 & 260.92 & 240.57 & 223.70 & 99.09\% & 99.17\% & 99.26\% & 99.19\% & 99.26\% & 99.30\% \\
    \hline
    Node+Edge, Pooling  & 469.48 & 392.53 & 357.46 & 344.27 & 321.45 & 300.77 & 98.92\% & 98.97\% & 99.00\% & 98.93\% & 98.99\% & 99.06\% \\
    \hline
    Edge                & 380.41 & 355.73 & 331.26 & 317.61 & 298.38 & 290.50 & 98.91\% & 98.99\% & 99.08\% & 98.85\% & 99.03\% & 99.15\% \\
    \hline
    Edge, Pooling       & 569.96 & 539.27 & 470.25 & 571.04 & 501.45 & 444.72 & 98.43\% & 98.51\% & 98.64\% & 98.20\% & 98.46\% & 98.64\% \\
\hline
\end{tabular}
\caption{\label{table:retraining_rates} Retraining Error and Prediction Rates}
\end{center}
\end{small}
\end{sidewaystable}

\begin{figure}
\centering
\begin{subfigure}[b]{0.49\textwidth}
\includegraphics[width=0.999\textwidth]{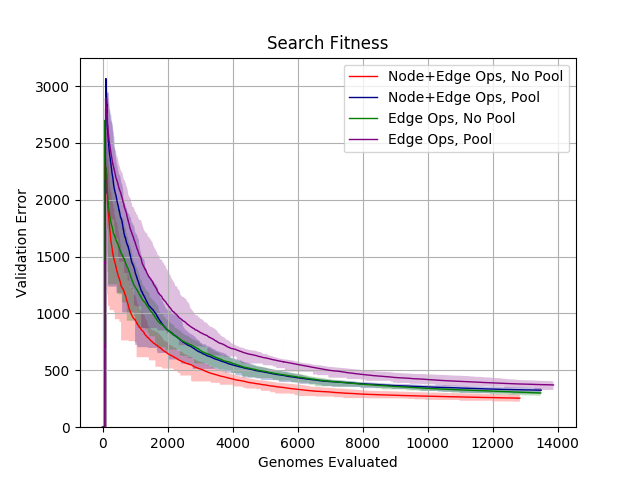}
\caption{\label{fig:search_progress} Fitness Progress}
\end{subfigure}
\begin{subfigure}[b]{0.49\textwidth}
\includegraphics[width=0.999\textwidth]{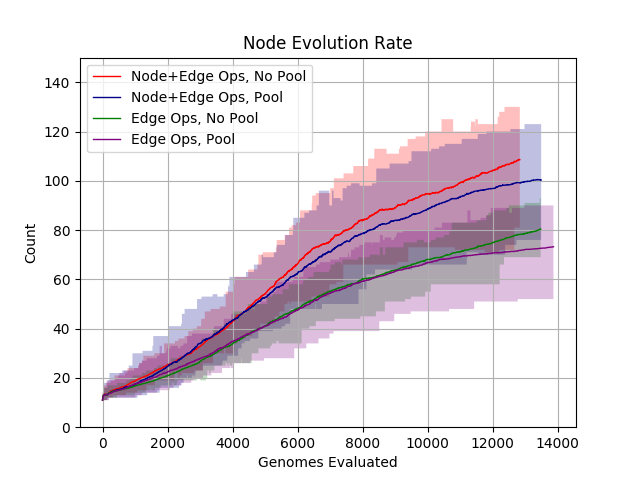}
\caption{\label{fig:node_progress} Node Counts}
\end{subfigure}
\begin{subfigure}[b]{0.49\textwidth}
\includegraphics[width=0.999\textwidth]{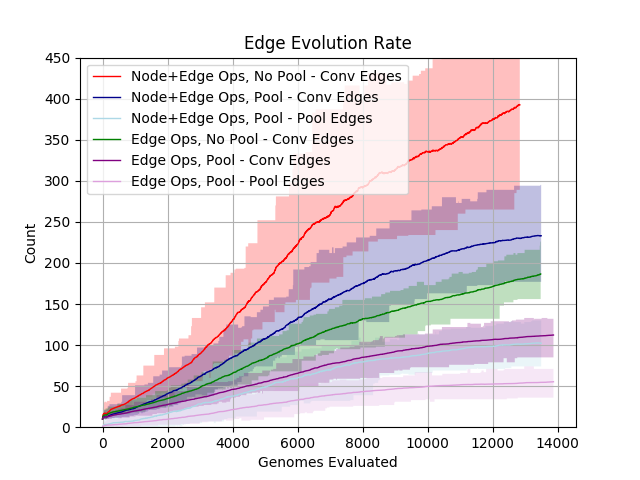}
\caption{\label{fig:edge_progress} Edge Counts}
\end{subfigure}
\begin{subfigure}[b]{0.49\textwidth}
\includegraphics[width=0.999\textwidth]{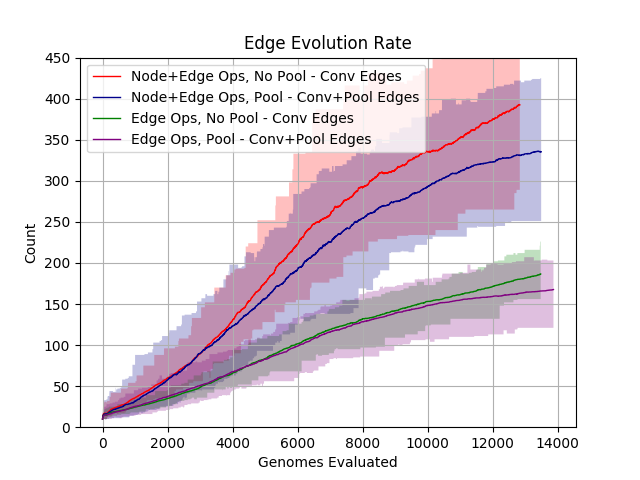}
\caption{\label{fig:total_edges_progress} Combined Edge Counts}
\end{subfigure}
\begin{subfigure}[b]{0.49\textwidth}
\includegraphics[width=0.999\textwidth]{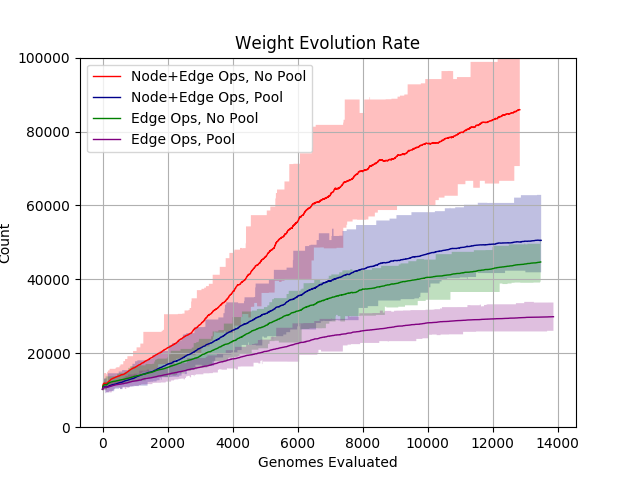}
\caption{\label{fig:weight_progress} Weight Counts}
\end{subfigure}
\caption{\label{fig:all_progress} These plots present the minimum, average maximum fitness and number of nodes, edges and weights across each type of search.}
\end{figure}

Figure~\ref{fig:all_progress} shows plots of the combined progress of each type of search, to provide an idea of the size and structure of the CNNs within the search populations as they progressed. The maximum and minimums are the maximum and minimum amounts across each search of that type, and the average is the average for all CNNs in each search of that type.

Figure~\ref{fig:search_progress} shows the range of CNN fitness (measured as validation error) for each search type. The searches with node mutations showed an improvement in both convergence rates and fitness of CNNs found, however, allowing for pooling edges hindered the search progress.  Figure~\ref{fig:node_progress} shows that in terms of the number of nodes in each genome, the pooling and non-pooling versions were fairly similar in the initial phases of the searches, however they began to diverge with the searches with node mutations starting to grow more rapidly around 5,000 evaluated CNNs, and the searches without node mutations diverging around 10,000 evaluated CNNs.  This suggests that allowing pooling potentially confounds the search space, preventing those searches from finding new improved CNNs.

Differences become even more apparent when examining the number of edges and trainable weights in the searches' CNNs. Figure~\ref{fig:edge_progress} shows the edge counts for CNNs within the search types, Figure~\ref{fig:total_edges_progress} shows the same information combining the pooling and convolutional edges, and Figure~\ref{fig:weight_progress} shows the weight counts. Node level mutations significantly increase the number of edges in the searches' genomes, while allowing pooling results in a decreased number of edges, and a even more dramatic decrease in number of weights.  This makes sense in that a pooling edge only has one weight for scaling, while a convolution contains weights equal to the filter size.  However, the reduced total number of edges may indicate that it is more challenging to find beneficial mutations when pooling edges are allowed.

\subsection{Mutation Operation Statistics}

\begin{figure}
    \begin{subfigure}[b]{\textwidth}
        \centering
        \includegraphics[width=0.49\textwidth]{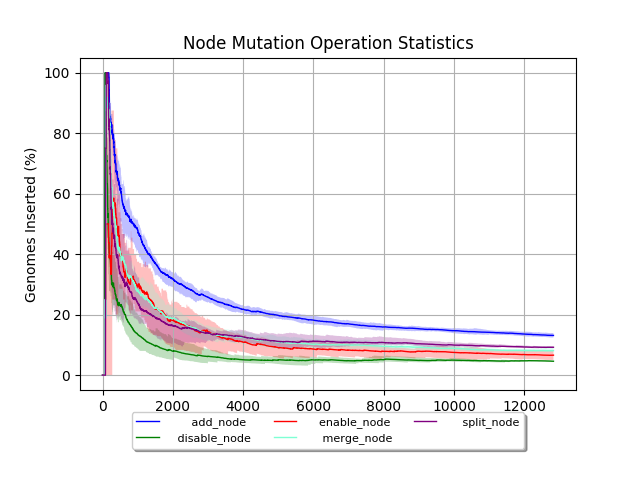}
        \includegraphics[width=0.49\textwidth]{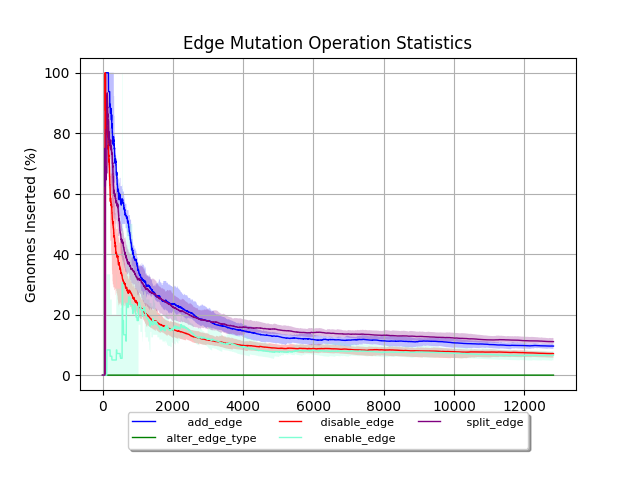}

        \includegraphics[width=0.49\textwidth]{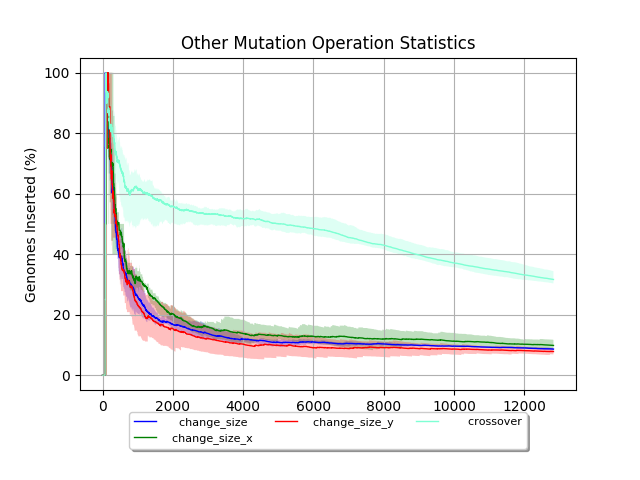}
        \caption{\label{fig:nodeops_nopool_operations} Nope+Edge ops. without pooling.}
    \end{subfigure}

    \begin{subfigure}[b]{\textwidth}
        \includegraphics[width=0.49\textwidth]{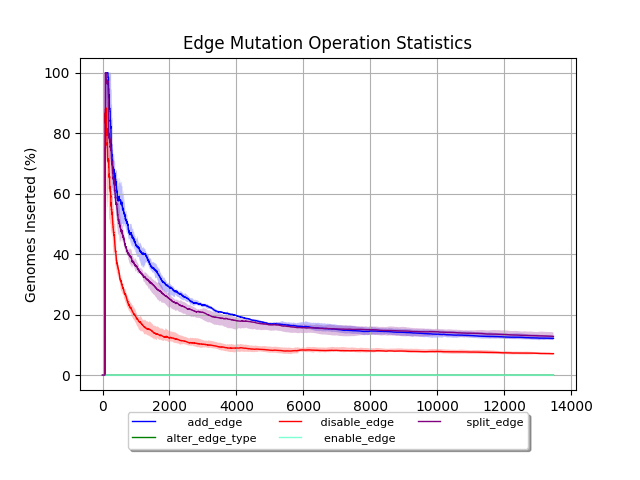}
        \includegraphics[width=0.49\textwidth]{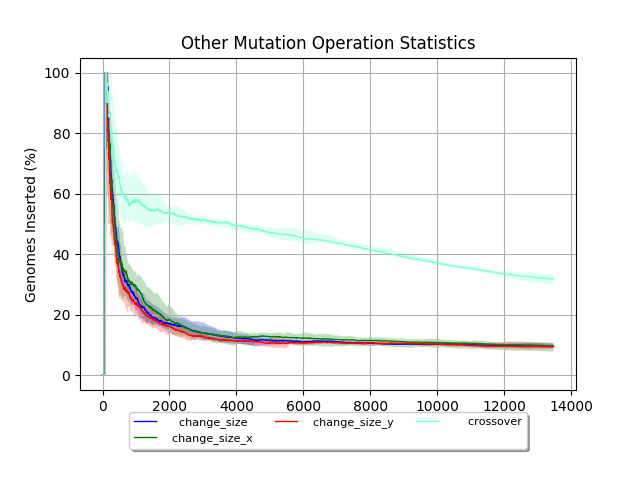}
        \caption{\label{fig:nopool_operations} Edge ops without pooling.}
    \end{subfigure}
    \caption{\label{fig:nopool_insertion_rates} Aggregated insertion rates for the search types without pooling.}
\end{figure}

\begin{figure}
    \begin{subfigure}[b]{\textwidth}
        \centering
        \includegraphics[width=0.49\textwidth]{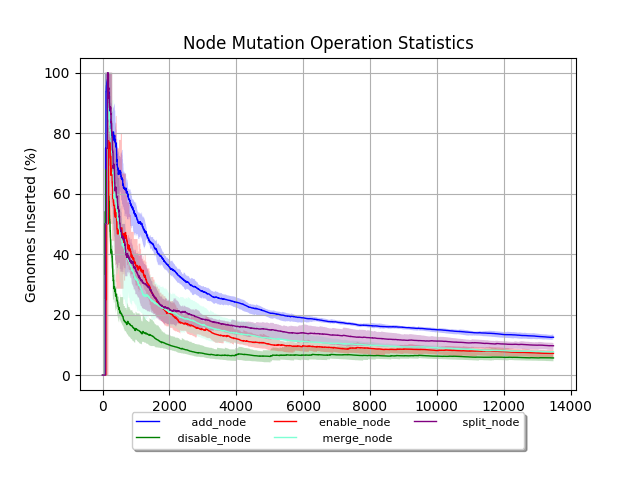}
        \includegraphics[width=0.49\textwidth]{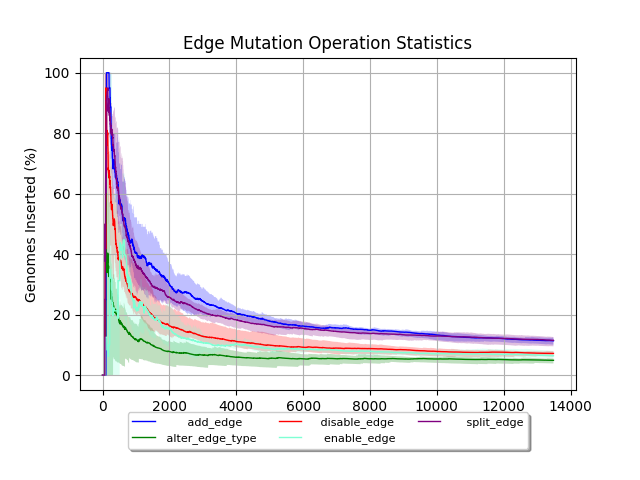}

        \includegraphics[width=0.49\textwidth]{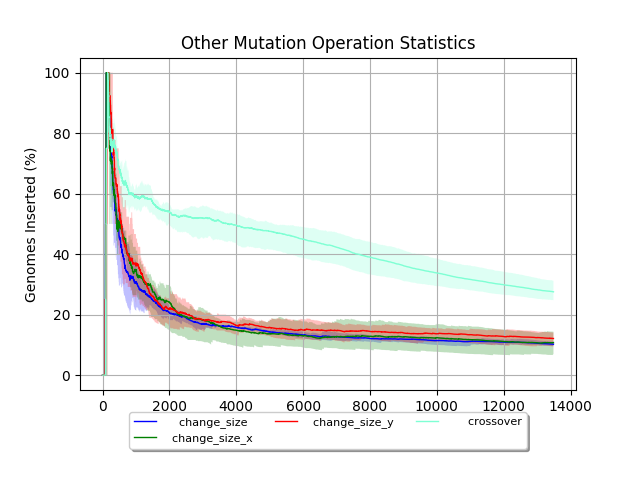}
        \caption{\label{fig:nodeops_pool_operations} Node+Edge ops. with pooling.}
    \end{subfigure}

    \begin{subfigure}[b]{\textwidth}
        \includegraphics[width=0.49\textwidth]{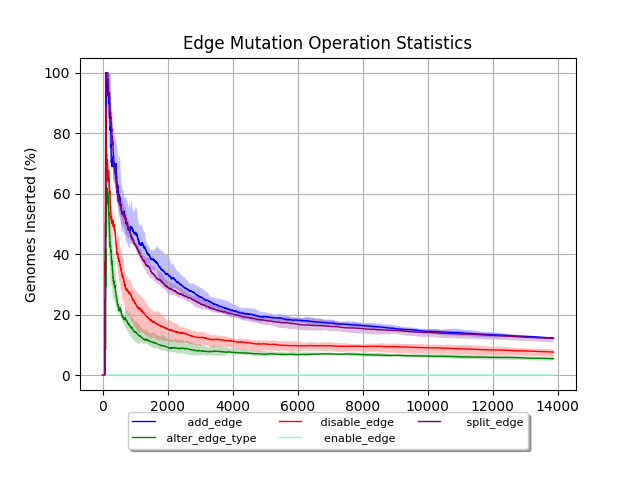}
        \includegraphics[width=0.49\textwidth]{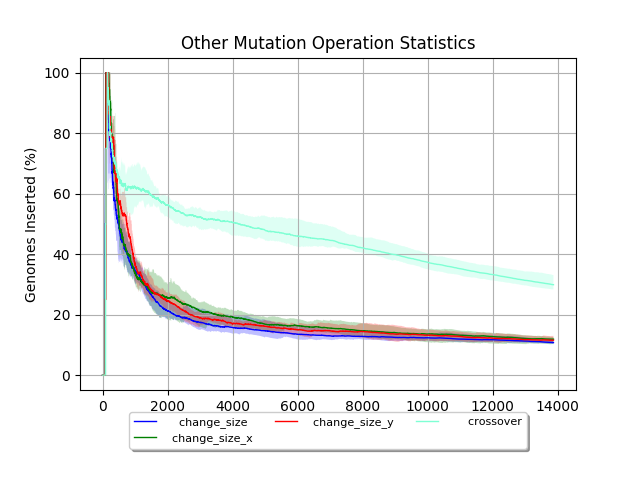}
        \caption{\label{fig:pool_operations} Edge ops. with pooling.}
    \end{subfigure}
    \caption{\label{fig:pool_insertion_rates} Aggregated insertion rates for the search types with pooling.}
\end{figure}

While each search was running, EXACT tracked what operator each new CNN was generated by and if it was inserted into the population.  For each search type, Figures~\ref{fig:nopool_insertion_rates} and~\ref{fig:pool_insertion_rates} present these insertion rates aggregated across the four searches. From these plots some interesting observations can be made.  The edge mutations, especially the {\emph add edge} and {\emph split edge} mutations, were more frequently inserted in searches with pooling enabled, both with and without node operations. The insertion rates for node operations are also slightly improved with pooling enabled. Conversely, pooling reduced the insertion rates from the crossover operator, which somewhat expectedly was the most successful operation as it combines components from two well structured and trained CNNs. There also was a wider variance of insertion rates in the searches with pooling enabled, which may be due to pooling increasing the search space as edges can be of two types.

This shows that pooling may provide some benefit to the mutation operations and performance of the CNNs, however allowing for pooling edges makes crossover a less effective operation which reduces the convergence rates of the searches and degrades overall performance.

\subsection{Epigenetic Weight Initialization}

Table ~\ref{table:best_nns} provides statistics for the validation and testing error and predictions of the best evolved CNNs from each of the four search types. Each of these CNNs were retrained with weights initialized by the standard strategy recommended by He \etal~\cite{he2015delving}, where weights, $w$, input to a neuron are initialized randomly with a variance, $\sigma^2(w) = \sqrt{\frac{2}{n}}$, where $n$ is the number of weights input to that neuron.  Table~\ref{table:retraining_rates} shows the min, average and maximum of these values, having retrained each of the best CNNs five times.

Even for the best retrained CNNs, test error rates were reduced by 0.11\% to 0.44\%, and on average by 0.20\% to 0.62\% which is quite significant as error rates were already under 1.0\%. Validation errors were closer, in the best case differences ranged between -0.06\% to 0.31\% and on average by 0.08\% to 0.21\%; which potentially indicates that epigenetic weight initialization helps train the CNNs to more {\it generalizable} sets of weights. These results are particularly interesting as, again, it should be mentioned that EXACT preserves CNNs in the populations utilizing the validation error rates, and the test data only used for final analyses -- the test data is never utilized in training the CNNs or by the EXACT algorithm, so its progress cannot be biased towards the test data.

\begin{table}
    \begin{center}
        \begin{small}
            \begin{tabular}{|l|r|r|r|r|r|}
                \hline
                {\bf Search}    & {\bf Weights} & {\bf Val.} & {\bf Test} & {\bf Val.} & {\bf Test} \\
                {\bf Type}      &               & {\bf Err.} & {\bf Err.} & {\bf Pred.} & {\bf Pred.} \\
                \hline
                \hline
                Node+Edge & 93813 & 259.49 & 188.62 & 99.20\% & 99.46\% \\
                Node+Edge, Pooling & 50387 & 335.53 & 271.21 & 98.99\% & 99.25\% \\
                Edge & 50285 & 292.04 & 225.64 & 99.12\% & 99.26\% \\
                Edge, Pooling & 30792 & 342.80 & 310.83 & 98.95\% & 99.08\% \\
                \hline
            \end{tabular}
            \caption{\label{table:best_nns} Best Evolved Neural Networks}
        \end{small}
    \end{center}
\end{table}

\subsection{Evolved Genomes}
\begin{figure*}
    \begin{subfigure}[b]{0.98\textwidth}
        \begin{center}
            \includegraphics[width=0.98\textwidth]{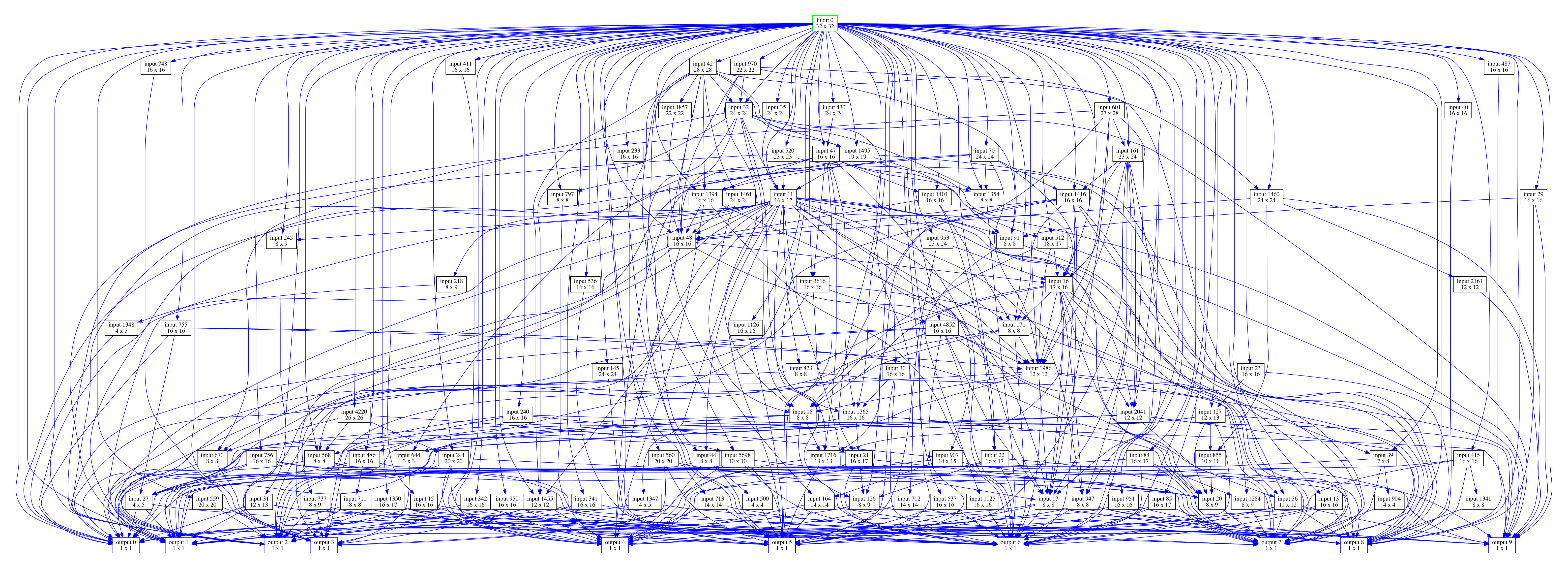}
            \caption{\label{fig:best_node_nn} This was the best CNN evolved by the searches using node and edge mutations and no pooling. It had a validation error of 259.488, test error of 188.616, validation accuracy of 99.20\% and test accuracy of 99.46\%. It has 112 nodes, 464 convolutional edges and 95,473 trainable weights.}
        \end{center}
    \end{subfigure}

    \begin{subfigure}[b]{0.98\textwidth}
        \begin{center}
            \includegraphics[width=0.98\textwidth]{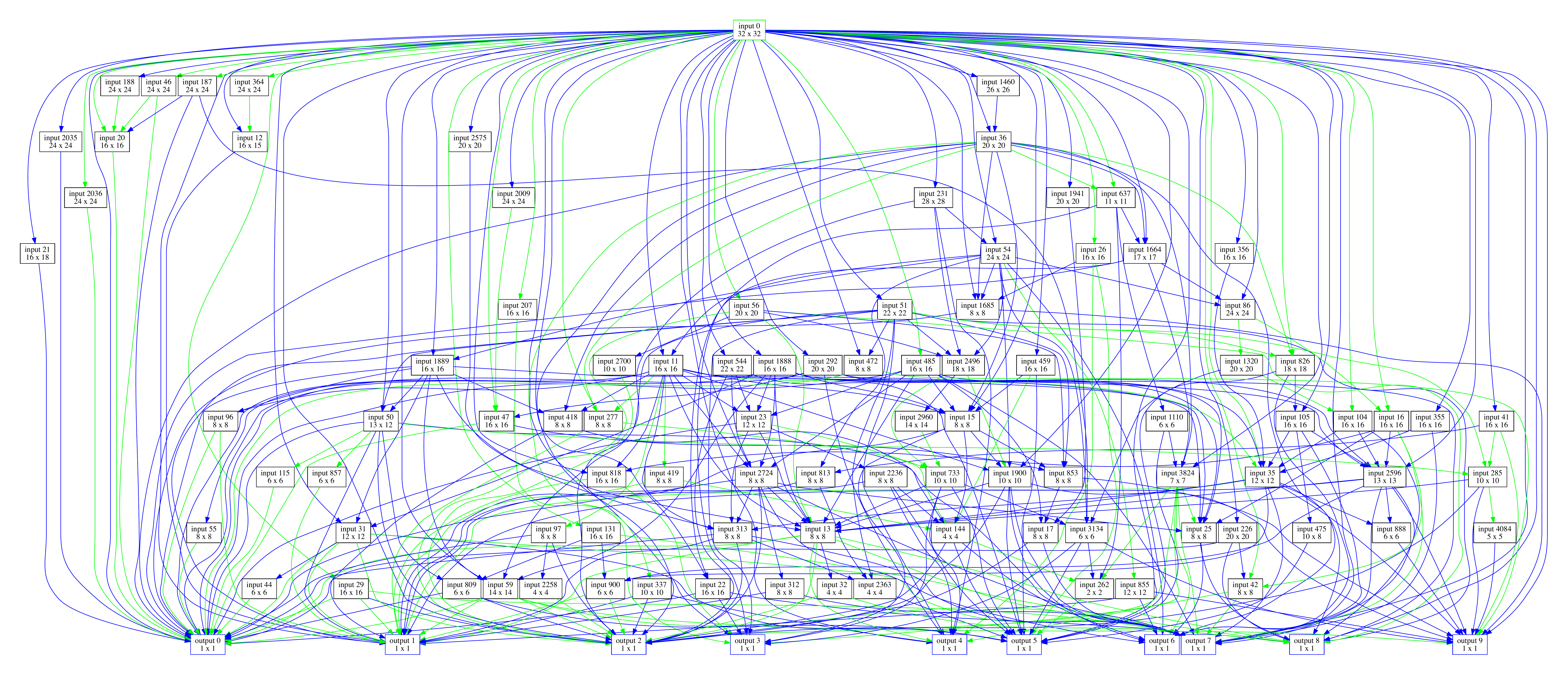}
            \caption{\label{fig:best_node_pool_nn} This was the best CNN evolved by the searches with pooling and using node and edge mutations. It had a validation error of 335.525, test error of 271.210, validation accuracy of 98.99\% and test accuracy of 99.25\%. It has 110 nodes, 285 convolutional edges, 116 pooling edges and 51,640 trainable weights.}
        \end{center}
    \end{subfigure}

    \begin{subfigure}[b]{0.98\textwidth}
        \begin{center}
            \includegraphics[width=0.98\textwidth]{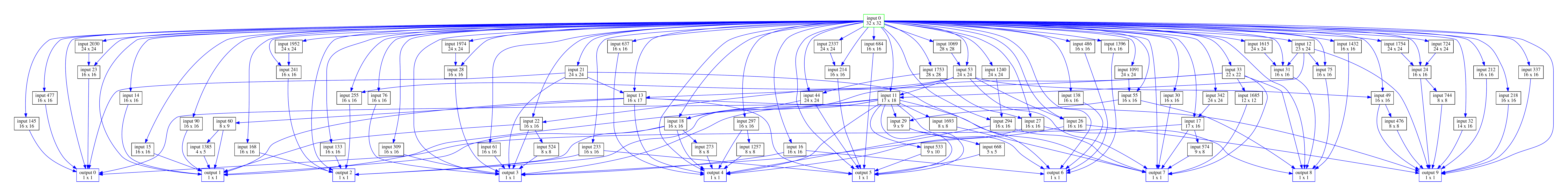}
            \caption{\label{fig:best_edge_nn} This was the best CNN evolved by the searches without pooling and using only edge mutations. It had a validation error of 292.035, test error of 225.635, validation accuracy of 99.12\% and test accuracy of 99.26\%. It has 91 nodes, 209 convolutional edges and 50,285 trainable weights.}
        \end{center}
    \end{subfigure}

    \begin{subfigure}[b]{0.98\textwidth}
        \begin{center}
            \includegraphics[width=0.98\textwidth]{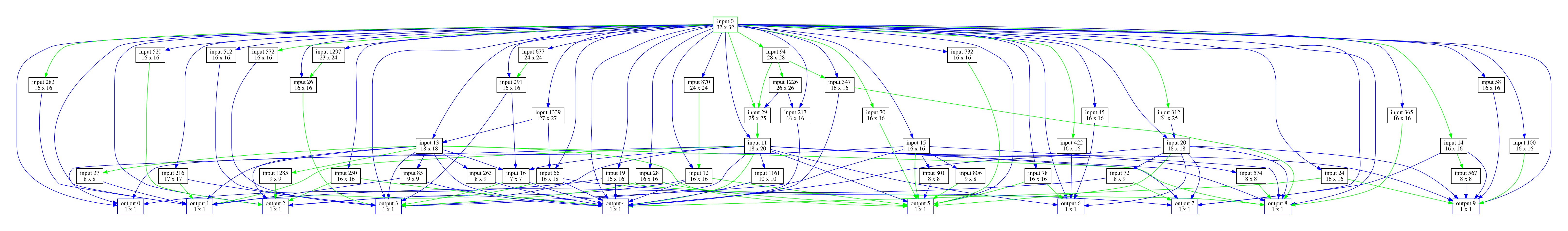}
            \caption{\label{fig:best_edge_pool_nn} This was the best CNN evolved by the searches with pooling and using only edge mutations. It had a validation error of 342.799, test error of 310.829, validation accuracy of 98.95\% and test accuracy of 99.08\%. It has 63 nodes, 108 convolutional edges, 50 pooling edges and 30,742 trainable weights.}
        \end{center}
    \end{subfigure}

    \caption{\label{fig:example_genomes} These four CNN genomes represent the CNNs evolved by the EXACT searches that performed best on the test data.}
\end{figure*}

Figure~\ref{fig:example_genomes} shows the best CNNs evolved by the searches. These networks are quite interesting in that they are highly different from the highly structured CNNs found seen in literature~\cite{lecun1998gradient,krizhevsky2012imagenet,simonyan2014very,szegedy2015going,he2016deep}. Some structures also appear to be forming, with some nodes being highly connected. Further, many connections skip layers, showing some similarity to ResNets~\cite{he2016deep}.

The best CNN, evolved by a search using node and edge mutations but no pooling reached a prediction accuracy of 99.46\% on the test data, with other similar searches not far behind -- results competitive with some of the best human designed CNNs~\cite{benenson_best_2017}. Further, for networks of this complexity only 13,000 to 14,500 evaluations may be low so there stands potential for further improvement. Figure~\ref{fig:all_progress} provides additional evidence that the CNNs have not finished evolving, as the number of nodes, edges and weights are still increasing without significant leveling off.  While the evolved CNNs are already quite accurate, it will be interesting to see if allowing for longer periods of evolution can result in more complicated network structures.

\section{Discussion and Future Work}
\label{sec:discussion}

This work presents an analysis of three different strategies for accelerating the neuro-evolution of convolutional neural networks. The first two, epigenetic weight initialization and node-level mutation operations are generic and can be applied to any neuro-evolution technique. Epigenetic weight initialization was shown to provide a significant improvement to test error rates, which were reduced by 0.11\% to 0.44\%, and on average by 0.20\% to 0.62\% when compared to weights initialized by the standard randomized method. Node mutations reduced test error by 0.21\% for the best found genomes and on average 0.175\% for the searches without pooling and by 0.18\% for the best found genomes and by 0.1575\% for the searches with pooling.  These results are significant as error rates were already below 1\%.

The third strategy, allowing edges to be either convolutional or fractional max pooling edges, was shown to degrade performance. While it was shown to slightly improve insertion rates for mutation operators, it reduced insertion rates for the crossover operator. Overall, it seems that the additional complexity of the search space caused by allowing for pooling edges did not overcome any benefit provided by allowing pooling.  This does, however, reflect on the sentiment of some members of the machine learning community, e.g. Geoffery Hinton describing the pooling operation as a ``big mistake''~\cite{hinton-reddit-ama} and other work showing well performing CNNs which do not utilize pooling~\cite{springenberg2014striving}.  That being said, the pooling operations used in this work did not allow for overlapping pools, which are shown to provide more benefit for many datasets~\cite{graham2014fractional}. Future work will investigate utilizing different overlaps on pooling edges.

%99.3775
%99.2025

%.175

%99.1925
%99.03

While these advances to the EXACT algorithm show significant potential for neuro-evolution of CNNs, due to computational requirements they have only been tested on the MNIST dataset. Current work is focusing on reproducing the same results on other data sets such as the CIFAR and TinyImage datasets~\cite{krizhevsky2009learning,torralba200880}. Other areas of interest involve performing multiple mutations when generating new CNNs, as a fair amount of time is taken for CNNs to reach sizes where they begin to become accurate. This may accelerate the evolution process even further.  Lastly, as EXACT can track insertion rates from the various recombination operators, additional benefits may be gained by online adaptation of how frequently the operations are used.  For example, it may improve performance to initial focus on operations which grow the CNN size (adding and splitting nodes and edges) and then later focus on operations which refine the CNN (disabling nodes and edges).

\bibliographystyle{abbrv}
\bibliography{./references}

\end{document}